\documentclass{article} 
\usepackage{iclr2026_conference,times}

\usepackage{amsmath,amsfonts,bm}

\def\eqref#1{equation~\ref{#1}}

\def\1{\bm{1}}

\DeclareMathAlphabet{\mathsfit}{\encodingdefault}{\sfdefault}{m}{sl}
\SetMathAlphabet{\mathsfit}{bold}{\encodingdefault}{\sfdefault}{bx}{n}

\usepackage{hyperref}
\usepackage{url}
\usepackage{lineno}
\usepackage{booktabs}       
\usepackage{amsfonts}       
\usepackage{nicefrac}       
\usepackage{microtype}      
\usepackage{lipsum}         
\usepackage{graphicx}
\usepackage{natbib}
\usepackage{doi}
\usepackage{enumitem}
\usepackage{multirow}
\usepackage{tabularx, booktabs}
\usepackage{amsthm}
\usepackage{hyperref}
\usepackage{fancyhdr}
\usepackage{amssymb}
\usepackage[table]{xcolor}
\usepackage{wrapfig}
\usepackage{fontawesome}
\usepackage{overpic}
\usepackage{subcaption}
\usepackage{tikz}
\usepackage[normalem]{ulem}

\nolinenumbers

\title{Improving Accuracy and Efficiency of Implicit Neural Representations: Making SIREN a \\WINNER}

\author{Hemanth Chandravamsi, $\quad$ Dhanush V. Shenoy, $\quad$ Steven H. Frankel \\
Faculty of Mechanical Engineering, Technion - Israel Institute of Technology, \\
Haifa, 3200003, Israel 
}

\iclrfinalcopy 
\begin{document}

\maketitle

\vspace{-15pt}
\begin{abstract}
We identify and address a fundamental limitation of sinusoidal representation networks (SIRENs), a class of implicit neural representations. SIRENs \cite{sitzmann2020implicit}, when not initialized appropriately, can struggle at fitting signals that fall outside their frequency support. In extreme cases, when the network’s frequency support misaligns with the target spectrum, a \textit{spectral bottleneck} phenomenon is observed, where the model yields to a near-zero output and fails to recover even the frequency components that are within its representational capacity. To overcome this, we propose WINNER - \underline{W}eight \underline{I}nitialization with \underline{N}oise for \underline{Ne}ural \underline{R}epresentations. WINNER perturbs uniformly initialized weights of base SIREN with Gaussian noise - whose noise scales are adaptively determined by the spectral centroid of the target signal. Similar to random Fourier embeddings, this mitigates `spectral bias' but without introducing additional trainable parameters. Our method achieves state-of-the-art audio fitting and significant gains in image and 3D shape fitting tasks over base SIREN. Beyond signal fitting, WINNER suggests new avenues in adaptive, target-aware initialization strategies for optimizing deep neural network training. For code and data visit \hyperlink{cfdlabtechnion.github.io/siren_square/}{cfdlabtechnion.github.io/siren\_square/}.
\end{abstract}

\section{Introduction} \label{sec:intro}
Implicit neural representations (INRs) enable the representation of coordinate-based discrete data such as images, videos, audio, 3D shapes, and physical fields as parameterized continuous functions. The key advantages of INRs are that they offer differentiable representations with efficient gradient evaluation and enable compressed representations suitable for optimization tasks across computer vision and scientific machine learning. Although conceptually simple, training deep networks for INRs can be challenging. Standard ReLU-based networks suffer from `spectral bias' - they tend to fit low frequencies while struggling to fit high-frequency components~\cite{rahaman2019spectral}. Approaches such as positional encodings and random Fourier features (RFFs) extend the dimensionality of input space to improve the frequency support and thus high-frequency learning~\cite{rahimi2007random,tancik2020fourier}. Alternatively, \cite{sitzmann2020implicit} introduced sinusoidal representation networks (SIRENs), which use periodic activations with a principled initialization.\\
\vspace{-12pt}

\begin{figure}[t!]
    \centering
    \includegraphics[width=\linewidth]{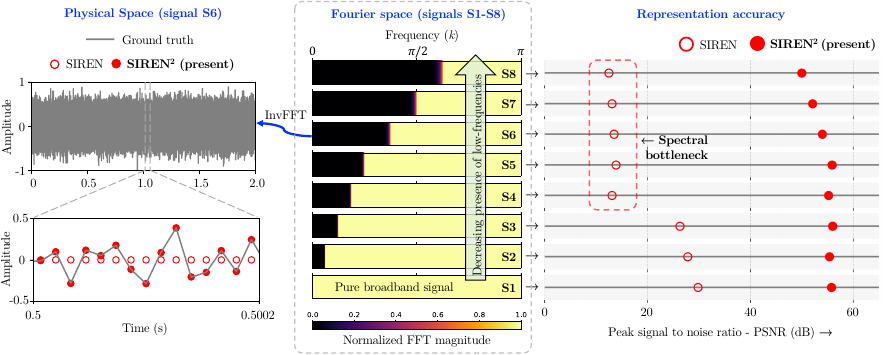}
    \vspace{-15pt}
    \caption{\textbf{Spectral bottleneck in SIRENs.} Eight broadband 1D signals S1–S8 (middle panel) are generated synthetically by masking the low-frequency components of pure broadband signal S1 to different extents. SIREN and SIREN$^2$ are supervised to fit these signals. The right panel shows PSNR values: performance of SIREN declines from S1 to S4 and collapses beyond S4, producing near-zero outputs. In contrast, SIREN$^2$, which uses the same architecture but with WINNER initialization, preserves accuracy for all signals. The left panel illustrates this for signal S6 in the time domain.}
    \label{fig:intro_fig}
\end{figure}

In a standard SIREN, the weights are initialized as $\mathbf{W}_{jk}^{(l)} \sim \mathcal{U}\!\left(-\tfrac{1}{\omega_0}\sqrt{\tfrac{6}{\text{fan\_in}}}, \tfrac{1}{\omega_0}\sqrt{\tfrac{6}{\text{fan\_in}}}\right)$, where $\omega_0$ is the sinusoidal activation frequency parameter. This initialization keeps pre-activations distributed in Gaussian with unit variance and maintains stabile training. However, reconstruction accuracy still depends strongly on the spectral composition of the target, and SIRENs often behave sensitive to $\omega_0$ and fail when the signal is dominated by very high or very low frequencies. As shown in Fig.~\ref{fig:intro_fig}, increasing the proportion of high frequencies in the target signal leads the network to collapse to a zero-valued output.  Prior studies~\cite{mehta2021modulated, liu2024finer} likewise report overfitting as signal length increases. While combining SIRENs with RFFs can mitigate this issue, it does so at the cost of a quadratic increase in parameter count with embedding size and network width. To avoid this overhead, we investigate \textit{weight initialization as a mechanism for modifying frequency support} without introducing additional parameters. We propose WINNER, a target-aware initialization scheme that \textit{adjusts the frequency domain properties of SIREN} at initialization to enable accurate representation of the target signal's frequency components. The results denoted as SIREN$^2$ in Fig.~\ref{fig:intro_fig} correspond to SIREN trained with WINNER initialization (\uline{SIREN + WINNER $\rightarrow$ SIREN$^2$}).

\section{Biased Learning Dynamics of SIREN} \label{sec:issue}
We show that SIRENs evolve to produce two distinct types of outputs (in terms spectral content) depending on the training target used for supervision. We compare two cases: a low-frequency 1D target, \texttt{bach.wav}, and a high-frequency 1D target, \texttt{tetris.wav}. Both signals are evaluated with a four-layer SIREN initialized using the standard scheme of ~\cite{sitzmann2020implicit} with inputs scaled to (-100,100). We examine the weight distributions and spectra of network's pre-activations at initialization and after training for $10^4$ epochs. Frequency response is quantified by the cumulative power spectral density of $N_h$ pre-activations per layer, $\mathrm{PSD}(k) = \sum_{j=1}^{N_h} |\widehat{x}_{\mathrm{pre},j}(k)|^2$. \\
\vspace{-12pt}

\textbf{Case 1, \texttt{bach.wav}:} While training with the low-frequency target, the initially narrow activation distributions of pre-activations become broader after training. Notable changes are observed in PSD between epoch 0 and epoch $10^4$ after layer-2. The network output preserves the dominant low-frequency components, and training converges towards the target without difficulty. \\
\vspace{-12pt}

\begin{figure}[h!]
    \centering
    \includegraphics[width=\linewidth]{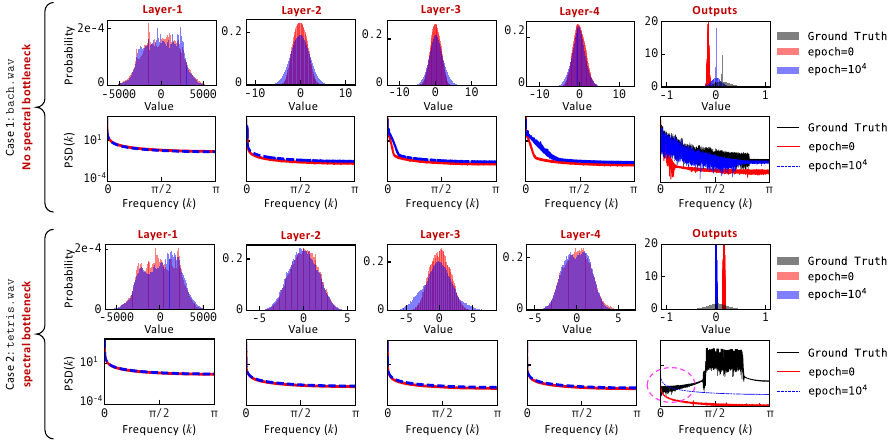}
    \vspace{-15pt}
    \caption{\textbf{Frequency support of SIREN for low- and high-frequency targets.} Shown are distributions (top row) and spectra (bottom row) of pre-activations from hidden layers 1–4 and the output when training on two audio signals, evaluated at initialization (epoch 0) and after $10^4$ epochs. \textit{Case 1 (low-frequency target, \texttt{bach.wav}):} the network maintains alignment with the target spectrum and avoids spectral bottleneck. \textit{Case 2 (high-frequency target, \texttt{tetris.wav}):} spectral energy remains concentrated at low frequencies, leading to a mismatch with the target and a spectral bottleneck.}
    \label{fig:bottleneck}
\end{figure}

\textbf{Case 2, \texttt{tetris.wav}:} The ground truth spectrum of \texttt{tetris.wav} differs substantially from spectral content of SIREN's pre-activations. This spectral mismatch leads to parameter updates that leave high-frequency content of output unaffected, preventing SIREN from learning \textit{even the low-frequency components of} \texttt{tetris.wav} that lie within its representational capacity. As shown in Fig.~\ref{fig:bottleneck} (last two rows), the distribution mass of outputs after $10^4$ epochs is concentrated near zero, and the resulting spectra does not align with the ground truth, even at a qualitative level. Essentially, the high-frequency content of \texttt{tetris.wav} \textit{acts as a bottleneck} the restricts the recovery of both high- and low-frequencies, a failure mode we term as the \emph{spectral bottleneck}. To overcome this, the spectral mismatch between the target and the network pre-activations must be reduced.\\
\vspace{-12pt}

\section{WINNER: A Target Aware Weight Initialization Scheme}
\begin{figure}[h!]
    \centering
    \includegraphics[width=\linewidth]{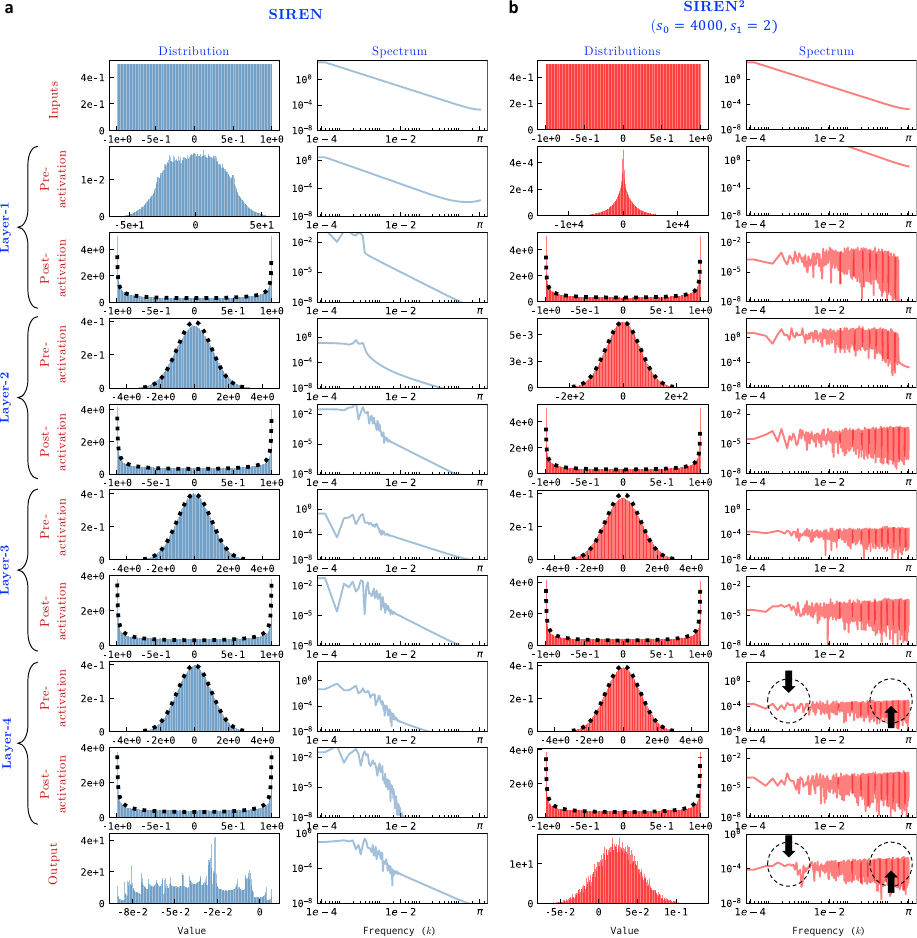}
    \vspace{-15pt}
    \caption{\textbf{Effect of WINNER initialization.} Shown are pre- and post-activation distributions and their power spectral densities at initialization for a four-layer SIREN (\textbf{a}) and SIREN$^2$ (\textbf{b}). In SIREN$^2$, Gaussian noise with scales $s_0$ and $s_1$ alters the pre- and post-activation spectra in the rightmost column. The circled regions indicate reduced energy at low frequencies accompanied by a corresponding increase at high frequencies. Analytical predictions shown in black dashed lines from \cite{chandravamsi2025spectral} and \cite{sitzmann2020implicit} align well with empirical results.}
    \label{fig:full_dists}
\end{figure}

To mitigate the spectral mismatch between the target and network pre-activations, and to address the spectral bottleneck shown in Figs.~\ref{fig:intro_fig} and \ref{fig:bottleneck}, we propose WINNER. Initialization schemes such as He or Glorot are designed to preserve the variance of activations across layers and avoid gradient explosion or decay, but in SIRENs this leads to smooth activation spectra that restrict frequency coverage. WINNER modifies this by perturbing the uniformly initialized weights $ \mathbf{W}_{jk}^{(l)} \sim \mathcal{U}\!\left(-\tfrac{1}{\omega_0}\sqrt{\tfrac{6}{\text{fan\_in}}}, \tfrac{1}{\omega_0}\sqrt{\tfrac{6}{\text{fan\_in}}}\right)$ in the first two layers by adding a Gaussian noise,  
\begin{equation}
    W_{jk}^{(l)} \leftarrow W_{jk}^{(l)} + \eta_{jk}^{(l)}, \quad 
    \eta_{jk}^{(l)} \sim \mathcal{N}\!\left(0, \tfrac{s}{\omega_0}\right), \quad
    s =
    \begin{cases}
        s_0, & l=1, \\
        s_1, & l=2, \\
        0, & l=3,...,L. \\
    \end{cases}
\end{equation}
This perturbation modifies the pre-activation distributions, increasing the standard deviation from $1$ (in the baseline SIREN) to $\sqrt{1 + \tfrac{d s^2}{2}}$, where $d$ is the input dimension and $s$ noise scale (see \cite{chandravamsi2025spectral} for proof). This also modifies the power-spectral density of pre-activations (Fig.~\ref{fig:bottleneck}), thereby enhancing the sensitivity of high-frequency target components to parameter updates during optimization. Based on the WINNER initialization, we define SIREN$^2$, a variant of SIREN with noise applied to the weights $\mathbf{W}^{(l)}$ upstream of first and second hidden layers:
\begin{equation}
f(\mathbf{x}; \theta) = \mathbf{W}^{(L)} \mathbf{h}^{(L-1)} + \mathbf{b}^{(L)}, \quad 
\mathbf{h}^{(l)} = 
\begin{cases}
\mathbf{x}, & l=0, \\
\phi^{\text{sin}}\!\big((\mathbf{W}^{(l)} + \eta^{(l)})\mathbf{h}^{(l-1)} + \mathbf{b}^{(l)}\big), & l=1,2, \\
\phi^{\text{sin}}\!\big(\mathbf{W}^{(l)}\mathbf{h}^{(l-1)} + \mathbf{b}^{(l)}\big), & l=3,\dots,L-1,
\end{cases}
\end{equation}
\begin{wrapfigure}{r}{0.3\textwidth}
    \vspace{-10pt}
    \begin{centering}
        \includegraphics[width=0.8\linewidth]{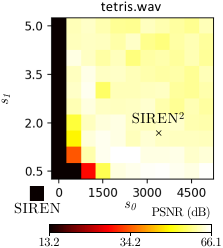}
    \end{centering}
    \caption{\textbf{Sensitivity to noise scales.} Heatmap of PSNR with different noise scales $s_0$ and $s_1$ tested on fitting the audio signal \texttt{tetris.wav}.}
    \label{fig:sense}
    \vspace{-5pt}
\end{wrapfigure}
where $\phi^{\text{sin}}=\sin{\omega_0 x}$ denotes the sinusoidal activation. Although noise is introduced only in the first two layers, its influence propagates downstream, altering the output spectrum and improving the ability to represent high-frequency content of the target. This effect is demonstrated in Fig.~\ref{fig:full_dists}.\\
\vspace{-12pt}

To adaptively set the noise scales, $s_0$ and $s_1$ are specified in a target-aware manner using the spectral centroid $\psi$, defined as
\[
    \psi = 2 \times \frac{\sum_k k\,|\hat{y}(k)|}{\sum_k |\hat{y}(k)|},
\]  
where $y(x)$ is the target signal, $\hat{y}(k)$ is its Fourier transform, and $k$ denotes frequency. Using $\psi$, the noise scales are computed as  
\begin{equation}
    s_0 = s_0^{\text{max}}\!\left(1 - e^{\,a\psi/C}\right), 
    \quad 
    s_1 = b\left(\tfrac{\psi}{C}\right),
\end{equation}
with \(C\) denoting the number of output channels. The hyperparameters \([s_0^{\text{max}}, a, b]\) are set to \([3500, 5, 3]\) for audio fitting and \([50, 5, 0.4]\) for image fitting. This formulation adapts initialization to the spectral profile of the target without introducing additional parameters. As shown in Fig.~\ref{fig:sense}, performance is robust across a wide range of $(s_0, s_1)$ values, and the target-aware rule yields near-optimal results while avoiding expensive hyperparameter searches.

\section{Summary}
We analyzed SIRENs under conditions where the spectrum of the target signal is strongly misaligned with that of the network pre-activations, such as when the target is dominated by high frequencies while the pre-activations have negligible spectral energy at high frequencies. In such cases, we identified and characterized a \textit{spectral bottleneck} phenomenon, where training collapses to near-zero outputs. To address this issue, we propose WINNER, an initialization strategy that broadens the frequency support of SIRENs by perturbing uniformly distributed weights with noise. Incorporating WINNER yields SIREN$^2$, a variant whose frequency support is adapted to the target. Appendix experiments on audio, images, and 3D fitting show PSNR improvements without additional trainable parameters, while denoising tasks benefit from tunable frequency support. Future work includes extending WINNER to other INR architectures and tasks, and developing more robust, generalized methods for estimating the noise scales $s_0$ and $s_1$.

\newpage
\section*{Acknowledgments}
The authors acknowledge the financial support provided by the Technion – Israel Institute of Technology during the course of this work.

\bibliography{iclr2026_conference}
\bibliographystyle{iclr2026_conference}

\newpage
\appendix
\section{Appendix}
\subsection{Related Work}
To address frequency-dependent limitations of SIRENs, recent approaches such as FINER \citep{liu2024finer} and FINER++ \citep{zhu2024finerplusplus} propose bias initialization schemes that mitigate eigenvalue decay in the empirical Neural Tangent Kernel and thereby extend the effective frequency support of the network. Complementary initialization strategies have been developed for sinusoidal and general INRs: VI$^3$NR derives an activation-agnostic, variance-preserving rule that stabilizes signal propagation across depth \citep{hewa2025vi3nr}; TUNER introduces a frequency-sampling input initialization together with amplitude-bounded hidden-layer weights to control the model bandlimit during training \citep{novello2025tuning}; and scaling initialization for sinusoidal neural fields accelerates optimization and alleviates spectral bias by multiplying all but the final layer’s weights by a constant factor \citep{yeom2024fast}. Other studies have examined how initialization strategies more generally influence optimization, including the work of \cite{tang2025structured} and \cite{varre2023spectral} in the setting of linear networks. In parallel, several model extensions have been introduced with the goal of improving INRs for audio data, including architectures for joint video–audio modeling, waveform encoding, and multi-resolution representations \citep{choudhury2024nerva,kim2023regression,kim2023generalizable,li2024asmr}.

\subsection{Experiments}
\textbf{Audio fitting:} We compare SIREN$^2$ (SIREN with WINNER initialization) to recent INR architectures, including the baseline SIREN~\cite{sitzmann2020implicit}, Gauss~\cite{ramasinghe2022beyond}, WIRE~\cite{saragadam2023wire}, and FINER~\cite{liu2024finer}, on a set of challenging audio reconstruction tasks. Each model is trained for 30{,}000 epochs, and results are averaged over five runs to report mean and standard-deviation of peak-PSNR. The dataset includes a mix of signals with low-, high-, and broadband spectral profiles. Table~\ref{tab:crit_fed} shows that SIREN$^2$ achieves higher reconstruction accuracy than other INR models, setting a new state-of-the-art on diverse audio signals. The method is especially effective for high-frequency targets, though it can under-represent low-frequency content (e.g., \texttt{bach.wav}) depending on the choice of $s_0$ and $s_1$.

\begin{table}[h!]
\centering
\caption{\textbf{Audio fitting.} Mean and standard deviation of PSNR scores for various architectures on audio reconstruction task. THe present SIREN$^2$ provides the best overall performance setting new \textit{state-of-the-art} results. Results are highlighted with \colorbox{red!30}{best}, \colorbox{green!30}{second best}, and \colorbox{cyan!30}{third best} reconstructions. Network widths are selected such that the total parameter count is close to the signal length.}
\vspace{-8pt}
\footnotesize
\begin{tabular}{l @{\hspace{0.01cm}} >{\centering\arraybackslash}p{1.7cm} @{\hspace{0.12cm}} >{\centering\arraybackslash}p{1.7cm} @{\hspace{0.12cm}} >{\centering\arraybackslash}p{1.7cm} @{\hspace{0.12cm}} > {\centering\arraybackslash}p{1.7cm} @{\hspace{0.12cm}} >{\centering\arraybackslash}p{1.7cm} @{\hspace{0.12cm}} >{\centering\arraybackslash}p{2.0cm}}
\toprule
& SIREN & FINER & WIRE & SIREN-RFF & FINER++ & \textbf{SIREN}$^\mathbf{2}$ \textbf{(present)} \\
\midrule
Hidden layers & \texttt{4$\times$222} & \texttt{4$\times$222} & \texttt{4$\times$157} &  \texttt{4$\times$193} & \texttt{4$\times$222} & \texttt{4$\times$222} \\
\# Fourier features & \texttt{0} & \texttt{0} & \texttt{0} & \texttt{193} & \texttt{0} & \texttt{0} \\
\# parameters & \texttt{149185} & \texttt{149185} & \texttt{149474} & \texttt{150155} & \texttt{149185} & \texttt{149185} \\
\midrule
\textbf{PSNR (dB)} ($\uparrow$):  \\[0.125cm]
\texttt{tetris.wav} & \texttt{13.4}$\pm$\texttt{0.0} & \texttt{13.6}$\pm$\texttt{0.0} & \texttt{13.6}$\pm$\texttt{0.0} & \colorbox{cyan!30}{\texttt{38.1}$\pm$\texttt{0.3}} & \colorbox{green!30}{\texttt{52.2}$\pm$\texttt{0.7}} & \colorbox{red!30}{\texttt{62.7}$\pm$\texttt{0.4}} \\
\texttt{tap.wav} & \texttt{20.4}$\pm$\texttt{0.0} & \texttt{21.1}$\pm$\texttt{0.0} & \texttt{21.1}$\pm$\texttt{0.0} & \colorbox{cyan!30}{\texttt{44.8}$\pm$\texttt{0.4}} & \colorbox{green!30}{\texttt{51.8}$\pm$\texttt{0.3}} & \colorbox{red!30}{\texttt{53.5}$\pm$\texttt{0.9}} \\
\texttt{whoosh.wav} & \texttt{33.8}$\pm$\texttt{0.9} & \colorbox{cyan!30}{\texttt{53.4}$\pm$\texttt{1.0}} & \texttt{20.2}$\pm$\texttt{0.0} & \texttt{41.8}$\pm$\texttt{0.6} & \colorbox{green!30}{\texttt{55.4}$\pm$\texttt{0.6}} & \colorbox{red!30}{\texttt{64.9}$\pm$\texttt{1.7}} \\
\texttt{arch.wav} & \texttt{29.7}$\pm$\texttt{1.1} & \colorbox{cyan!30}{\texttt{58.5}$\pm$\texttt{0.8}} & \texttt{17.2}$\pm$\texttt{0.1} & \texttt{44.1}$\pm$\texttt{0.9} & \colorbox{green!30}{\texttt{65.2}$\pm$\texttt{0.2}} & \colorbox{red!30}{\texttt{95.2}$\pm$\texttt{2.9}} \\
\texttt{relay.wav} & \texttt{28.5}$\pm$\texttt{1.4} & \texttt{34.7}$\pm$\texttt{0.5} & \texttt{20.7}$\pm$\texttt{0.0} & \colorbox{cyan!30}{\texttt{40.5}$\pm$\texttt{0.6}} & \colorbox{green!30}{\texttt{54.1}$\pm$\texttt{0.4}} & \colorbox{red!30}{\texttt{60.4}$\pm$\texttt{2.9}} \\
\texttt{voltage.wav} & \texttt{34.0}$\pm$\texttt{0.8} & \colorbox{cyan!30}{\texttt{53.4}$\pm$\texttt{0.6}} & \texttt{20.0}$\pm$\texttt{0.0} & \texttt{43.7}$\pm$\texttt{0.3} & \colorbox{green!30}{\texttt{56.5}$\pm$\texttt{0.1}} & \colorbox{red!30}{\texttt{64.5}$\pm$\texttt{0.5}} \\
\texttt{bach.wav} & \texttt{59.4}$\pm$\texttt{0.3} & \colorbox{red!30}{\texttt{64.5}$\pm$\texttt{0.2}} & \texttt{26.1}$\pm$\texttt{0.5} & \texttt{41.8}$\pm$\texttt{0.2} & \colorbox{green!30}{\texttt{62.2}$\pm$\texttt{0.3}} & \colorbox{cyan!30}{\texttt{60.5}$\pm$\texttt{0.2}} \\
\midrule
Average & \texttt{31.3}$\pm$\texttt{0.6} & \texttt{42.7}$\pm$\texttt{0.5} & \texttt{19.9}$\pm$\texttt{0.1} & \texttt{42.1}$\pm$\texttt{0.5} & \texttt{56.8}$\pm$\texttt{0.4} & \texttt{66.0}$\pm$\texttt{1.4} \\
\bottomrule
\end{tabular}
\label{tab:crit_fed}
\end{table}

\begin{figure}
    \centering
    \includegraphics[width=\linewidth]{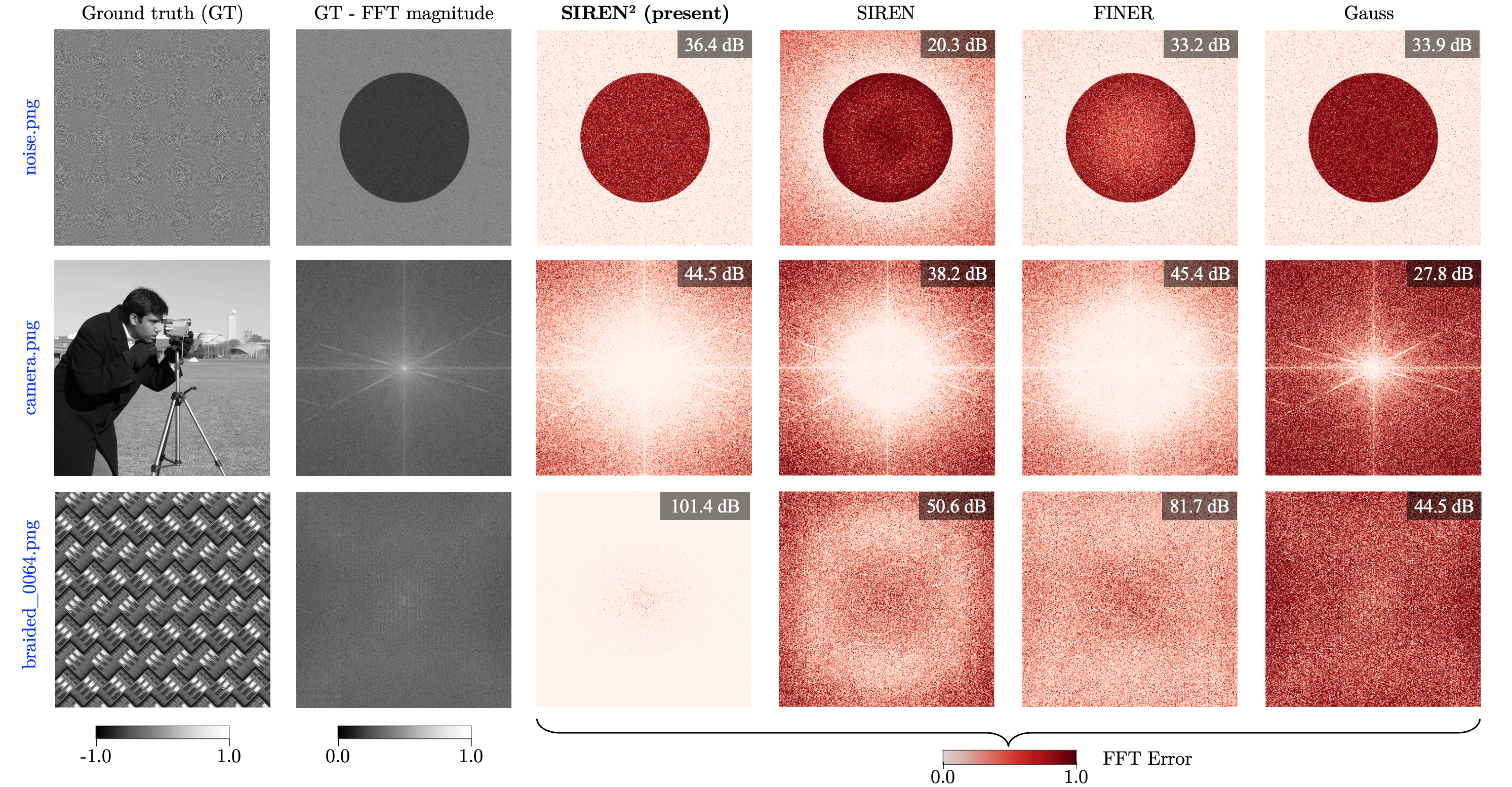}
    \caption{\textbf{Accuracy improvements in Fourier space with SIREN$^2$ for image fitting.} Reconstruction performance of SIREN, SIREN$^2$, FINER, and Gauss on different images. The first two columns show the ground truth image samples and their FFT magnitudes. The following columns show FFT error maps along with peak PSNR values obtained with different models. SIREN$^2$ consistently yields lower FFT errors and higher PSNR compared to the baseline SIREN.}
    \label{fig:image_fitting}
\end{figure}

\textbf{2D Image fitting:} We evaluate SIREN$^2$ on 2D image fitting, $f(\mathbf{x};\theta): \mathbb{R}^2 \mapsto \mathbb{R}^d$, with $d=1$ for grayscale and $d=3$ for RGB. Datasets include natural images (\cite{kodak_photocd}), texture data from DTD~\cite{cimpoi14describing} (\emph{braided}), and a synthetic high-frequency noisy image (\texttt{noise.png}). As shown in Table~\ref{tab:images1}, SIREN$^2$ consistently improves over SIREN, with PSNR gains from 9\% (\texttt{castle.jpg}) to 69\% (\texttt{noise.png}). The gains with SIREN$^2$ are particularly notable for images with high-frequency content and low pixel counts (i.e., more parameters per pixel). For images with high pixel counts, the improvements are marginal unless the spectrum is dominated by high frequencies; in such cases, FINER yields the best PSNR. Figure~\ref{fig:image_fitting} and corresponding FFT error maps show that SIREN$^2$ reduces high-frequency fitting errors and preserves fine details better then SIREN consistently.

\begin{table}[h!]
\centering
\caption{\textbf{2D image fitting.} Peak PSNR ($\uparrow$) in dB for different images and datasets across INR models. SIREN$^2$ surpasses the baseline SIREN, with percentage gains (in parentheses) arising solely from initialization. Reconstructions are highlighted as \colorbox{red!30}{best}, \colorbox{green!30}{second best}, and \colorbox{cyan!30}{third best}.}
\vspace{-8pt}
\scriptsize
\begin{tabular}{l @{\hspace{0.01cm}} >{\centering\arraybackslash}p{1.0cm} @{\hspace{0.06cm}} >{\centering\arraybackslash}p{1.6cm} @{\hspace{0.06cm}} >{\centering\arraybackslash}p{1.0cm} @{\hspace{0.06cm}} > {\centering\arraybackslash}p{1.0cm} @{\hspace{0.06cm}} > {\centering\arraybackslash}p{1.0cm} @{\hspace{0.06cm}} >{\centering\arraybackslash}p{1.0cm}}
\toprule
& SIREN & \textbf{SIREN$^2$ (present)} & ReLU-PE & WIRE & FINER & Gauss \\
\midrule
Hidden layers ($n\times w$) & \texttt{4$\times$256} & \texttt{4$\times$256} & \texttt{4$\times$256} & \texttt{4$\times$128} & \texttt{4$\times$256} & \texttt{4$\times$256} \\
\# parameters & \texttt{198145} & \texttt{198145} & \texttt{263553} & \texttt{198386} & \texttt{198145} & \texttt{198145}  \\
\midrule
\textbf{Peak PSNR (dB)} ($\uparrow$):  \\[0.125cm]
noise.png & 21.3 & \colorbox{red!30}{\textbf{36.1}} (\textcolor{teal}{69\% $\uparrow$}) & 16.9 & 25.5 & \colorbox{cyan!30}{33.0} & \colorbox{green!30}{34.1} \\
camera.png & \colorbox{cyan!30}{38.9} & \colorbox{green!30}{\textbf{44.9}} (\textcolor{teal}{15\% $\uparrow$}) & 28.4 & 37.2 & \colorbox{red!30}{46.4} & 28.6 \\
castle.jpg & \colorbox{cyan!30}{33.6} & \colorbox{green!30}{\textbf{36.5}} (\textcolor{teal}{9\% $\uparrow$}) & 22.3 & 28.5 & \colorbox{red!30}{36.9} & 19.2 \\
DTD braided dataset (120 images, gray mode)$^*$ & \colorbox{cyan!30}{48.6} & \colorbox{red!30}{\textbf{75.2}} (\textcolor{teal}{55\% $\uparrow$}) & - & - & \colorbox{green!30}{65.4} & - \\
Kodak dataset (24 images, gray mode)$^*$ & \colorbox{cyan!30}{34.9} & \colorbox{green!30}{\textbf{37.6}} (\textcolor{teal}{8\% $\uparrow$}) & - & - & \colorbox{red!30}{38.1} & - \\
\bottomrule
\multicolumn{7}{r}{\scriptsize $^*$The reported PSNR values for these datasets represent averages computed over all images.}
\end{tabular}
\label{tab:images1}
\end{table}

\begin{figure}
    \centering
    \includegraphics[width=\linewidth]{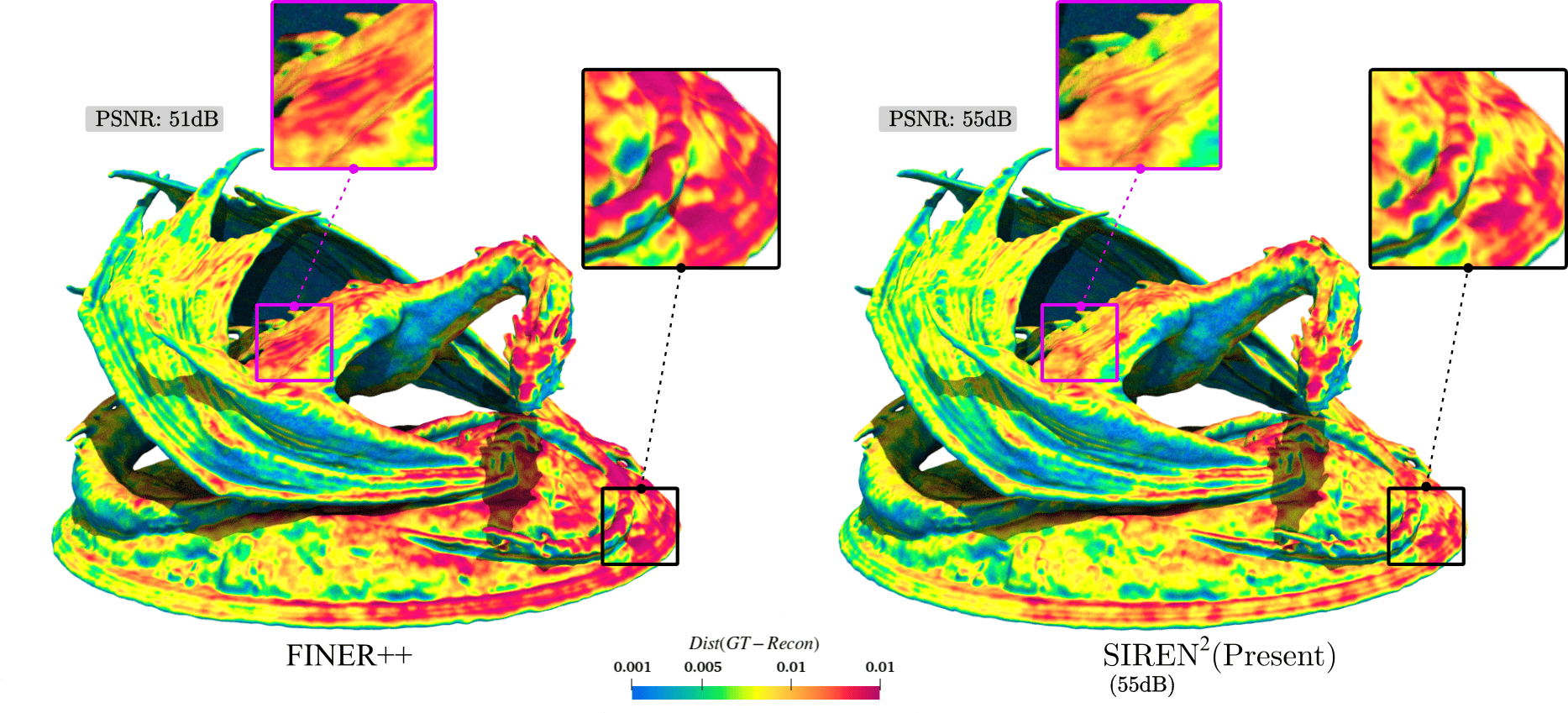}
    \caption{Comparison of FINER++ and SIREN$^2$ in fitting signed distance functions (SDFs) from oriented point clouds. Both approaches yield high-quality reconstructions, with SIREN$^2$ showing slightly superior performance. Baseline SIREN, WIRE, and FINER gave considerably worse results and are therefore omitted.}
    \label{fig:dragon_error}
\end{figure}

\textbf{3D shape fitting:} We fit signed distance functions (SDFs) to oriented point clouds adopting the framework of Sitzmann et al.~\cite{sitzmann2020implicit} with a composite loss that combines SDF, Eikonal, normal, and far/outside terms, following recent SDF reconstruction methods~\cite{peng2023gens_3D, atzmon2020sal_3D}. For each batch, 3D points are sampled near the surface and in the surrounding volume, and the network predicts signed distances and normals. Both FINER++ and SIREN$^2$ are trained for 10,000 epochs; reconstructions reach peak PSNR of $51$ dB and $55$ dB, respectively. Geometric fidelity is evaluated against the ground-truth (Fig.~\ref{fig:dragon_error}), where SIREN$^2$ exhibits lower vertex-wise error, especially in regions with fine-scale curvature and high-frequency detail. While FINER++ oversmooths sharp features and introduces local distortions, SIREN$^2$ better preserves geometric structure with tighter surface alignment.

\begin{figure}[t!]
    \centering
    \includegraphics[width=\linewidth]{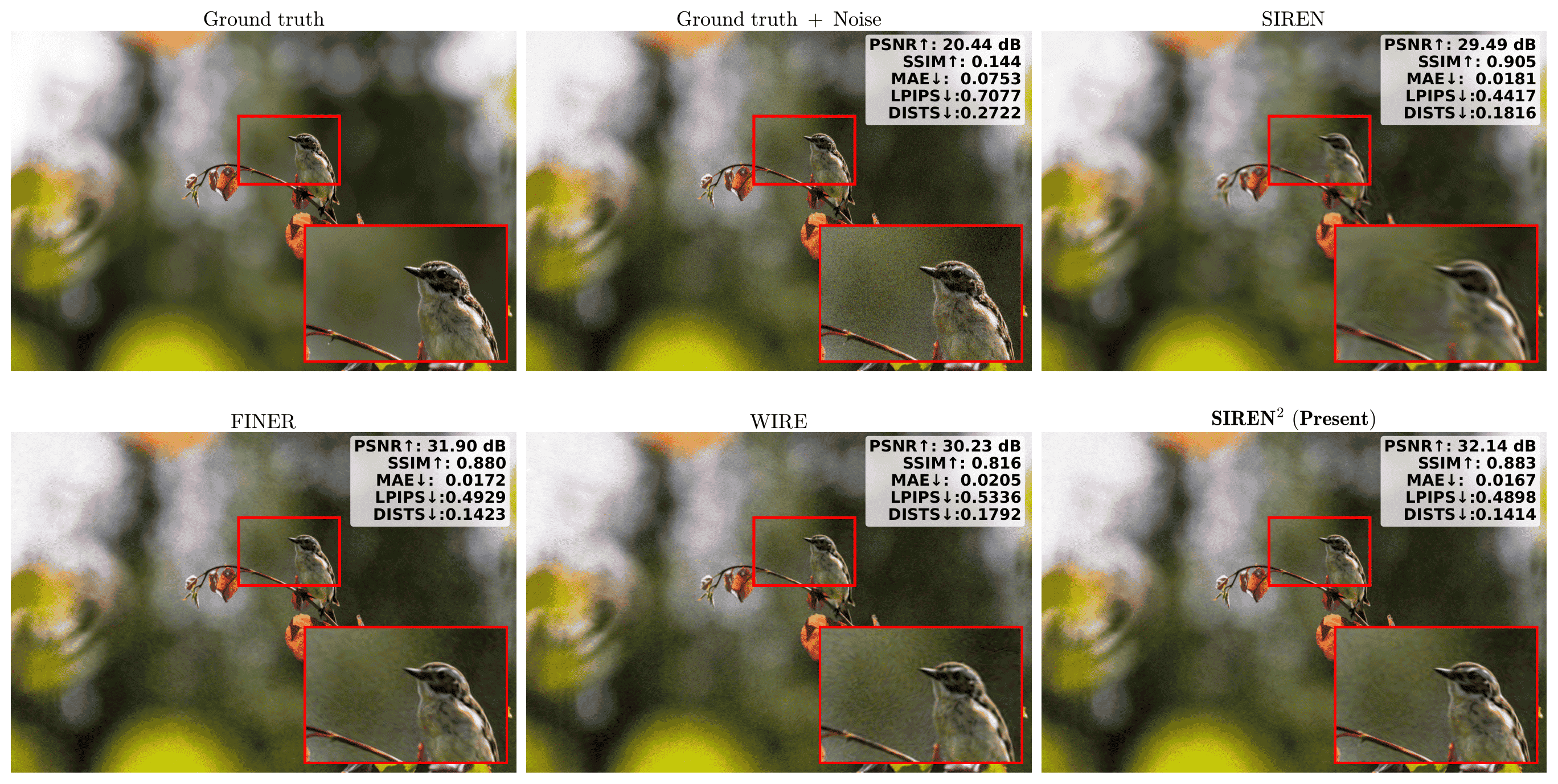}
    \caption{\textbf{Image denoising.} Denoising results of a sparrow image using different network architectures. The insets highlight regions of interest, illustrating the improved preservation of fine details and overall denoising performance of SIREN$^2$.}
    \label{fig:image-denoise}
\end{figure}

\textbf{Image denoising:} The robustness of INR architectures is evaluated on canonical image denoising, where a clean signal $f(\mathbf{x})$ is corrupted with additive white Gaussian noise $\eta(\mathbf{x}) \sim \mathcal{N}(0,\sigma^2)$ to yield $\tilde{f}(\mathbf{x}) = f(\mathbf{x}) + \eta(\mathbf{x})$ at $\mathrm{SNR}=5\ \mathrm{dB}$. The task is to recover $f$ from $\tilde{f}$ using an unsupervised strategy similar to Noise2Self~\cite{batson2019noise2self}, training directly on noisy inputs while holding out a small pixel subset for J-invariant validation, which prevents pixelwise memorization and provides an unsupervised early-stopping criterion. Ground-truth images are used \textit{only for evaluation} (PSNR, SSIM, MAE, LPIPS~\cite{zhang2018unreasonable}, and DIST~\cite{ding2020image}), with arrows ($\uparrow$/$\downarrow$) indicating metric direction. As shown in Fig.~\ref{fig:image-denoise}, SIREN and FINER oversmooth fine-scale details, WIRE retains global appearance but leaves residual noise, while SIREN$^2$ yields the sharpest and most faithful reconstructions with the best SSIM and LPIPS, suppressing noise while preserving textures and edges.

\begin{figure}[h!]
    \centering
    \includegraphics[width=\linewidth]{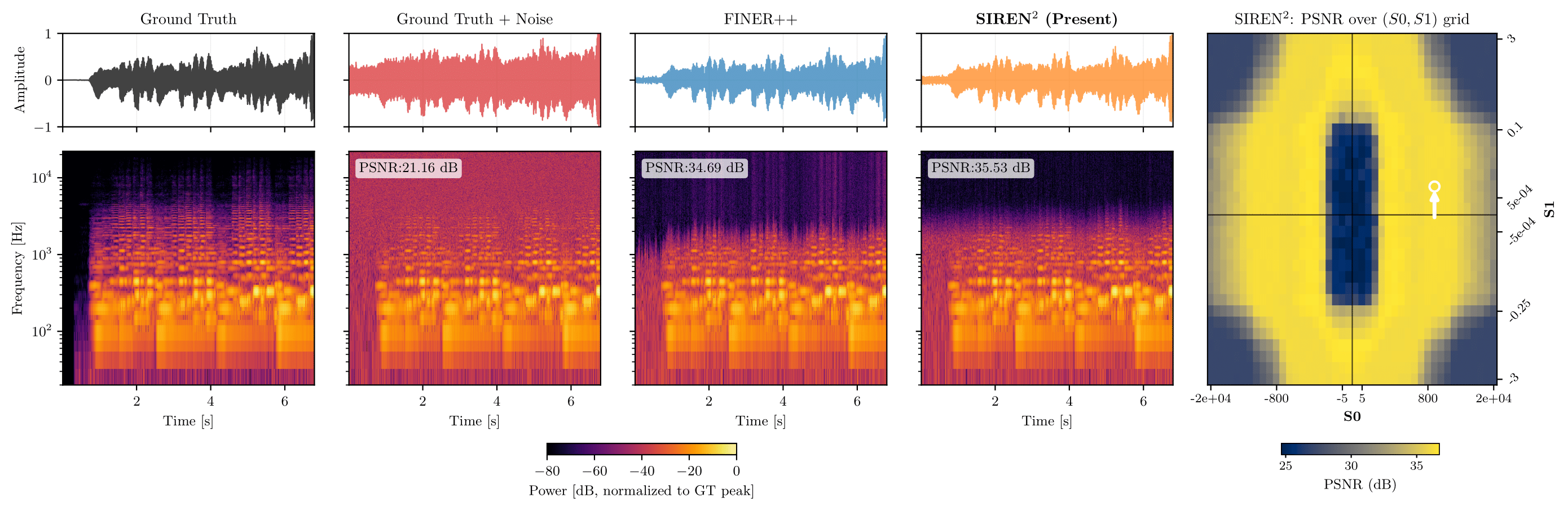}
    \vspace{-20pt}
    \caption{\textbf{Audio denoising.} Log-spectrogram comparison for the Bach audio denoising task. The ground-truth signal (top left) is corrupted with additive Gaussian noise at $5\,\text{dB}$ SNR. Reconstructions using SIREN$^2$ recover the smooth structures of the signal effectively.}
    \label{fig:audio-denoise}
\end{figure}

\begin{wraptable}{r}{0.40\textwidth}
\centering
\caption{\textbf{Audio denoising.} Best PSNR ($\uparrow$) in dB for different audio clips using FINER++ and SIREN$^2$.}
\vspace{-5pt}
\scriptsize
\begin{tabular}{l 
  @{\hspace{0.01cm}} >{\centering\arraybackslash}p{1.1cm}
  @{\hspace{0.01cm}} >{\centering\arraybackslash}p{1.1cm} 
  @{\hspace{0.01cm}} >{\centering\arraybackslash}p{1.1cm} }
\toprule
 & $\enspace$ GT + $\quad$ Noise & FINER++ & \textbf{SIREN$^2$ (present)} \\
\midrule
\texttt{bach.wav}      & 21.16 & 34.69 & \textbf{35.53} \\
\texttt{dilse.wav}     & 21.46 & 35.16 & \textbf{35.18} \\
\texttt{birds.wav}     & 29.92 & 34.84 & \textbf{37.17} \\
\texttt{counting.wav}  & 26.67 & 38.19 & \textbf{38.71} \\
\bottomrule
\end{tabular}
\vspace{-10pt}
\label{tab:audio-denoise}
\end{wraptable}
\textbf{Audio denoising:} We adopt the same Noise2Self-inspired DIP setup~\cite{batson2019noise2self} as in the image denoising experiments, with ground-truth signals used solely for evaluation. Unlike images, where most energy is typically in low frequencies, natural and synthetic audio signals often exhibit broadband structure (e.g., harmonics, ambient noise) that overlaps with Gaussian noise and limits frequency-based filtering. In supervised audio fitting, all models used $\omega_0$ scaling by 100, but this degrades denoising performance ($\mathrm{PSNR}<25\,\mathrm{dB}$) due to broadband noise. We therefore omit first-layer $\omega_0$ scaling used in the audio fitting task. Without the first-layer scaling only FINER++ and SIREN$^2$ provided satisfactory results in our experiments; so we only present results from these two architectures. As shown in Table~\ref{tab:audio-denoise} and Fig.~\ref{fig:audio-denoise}, both achieve competitive denoising accuracy on \texttt{bach.wav} after 20,000 epochs. In SIREN$^2$, the noise scales $s_0$ and $s_1$ act as tunable filters for adjusting frequency support, while FINER++ performance varies with bias initialization.

\subsection*{Reproducibility Details}
For detailed reproducibility information on all experiments presented in this paper, please refer to our main paper \cite{chandravamsi2025spectral}. The code and data used in this work are available at \hyperlink{cfdlabtechnion.github.io/siren_square/}{cfdlabtechnion.github.io/siren\_square/}.

\end{document}